%% file: HITL_paper.tex
\documentclass{article}
\usepackage{caption}
\usepackage{microtype}
\usepackage{graphicx}
\usepackage{subfigure}
\usepackage{booktabs} 
\usepackage{todonotes}
\presetkeys{todonotes}{inline}{}

\usepackage{amsmath}
\usepackage{amsfonts}
\usepackage{amssymb}
\usepackage{calligra}
\usepackage{calrsfs}

\newcommand{\context}{\mathcal{C}}
\newcommand{\cluster}{\mathbb{C}}
\newcommand{\embedding}{E}
\newcommand{\candidate}{c}
\newcommand{\pos}{+}
\newcommand{\negs}{-}
\newcommand{\anomaly}{a}

\usepackage{hyperref}

\usepackage[accepted]{icml2020}

\icmltitlerunning{Submission and Formatting Instructions for ICML 2020}

\begin{document}

\twocolumn[
\icmltitle{Improve black-box sequential anomaly detector relevancy with limited user feedback}




\begin{icmlauthorlist}
\icmlauthor{Luyang Kong}{aws}
\icmlauthor{Lifan Chen}{aws}
\icmlauthor{Ming Chen}{aws}
\icmlauthor{Parminder Bhatia}{aws}
\icmlauthor{Laurent Callot}{aws}
\end{icmlauthorlist}

\icmlaffiliation{aws}{Amazon Research}

\icmlcorrespondingauthor{Laurent Callot}{lcallot@amazon.com}

\icmlkeywords{Anomaly Detection, Human In The Loop, ICML}

\vskip 0.3in
]



\printAffiliationsAndNotice{} 

\begin{abstract}

\input{abstract}
\end{abstract}

\section{INTRODUCTION}
\label{problem_statement}
\input{introduction}

\section{RELATED WORK}
\input{related_work}

\section{METHODOLOGY}

\input{method3}

\section{DATASETS}
\input{dataset}

\medskip
\section{EXPERIMENTS}
\input{experiment}

\section{Conclusion}


In this paper, we propose a general method to improve the relevancy of the anomalies generated by any black-box statistical anomaly detector based on sparse user feedback. The method employs the clustering of anomalies in an embedding space and computes a relevancy score based on the concentrations of negative and positive feedback within each cluster. With limited number of feedback required, empirical results show the method improves the precision and recall over a range of different anomaly detectors significantly on both synthetic and real-world datasets.

\section*{Acknowledgements}
We thank all the team members in AWS AI Lab for meaningful discussions. We thank Tim Januschowski, Bing Xiang and Taha Kass-Hout for organizing the collaboration. Moreover, we thank Kristjan Arumae for providing valuable feedbacks. 


\bibliography{HITL_paper}
\bibliographystyle{icml2020}
\nocite{}

\end{document}

%% file: abstract.tex
Anomaly detectors are often designed to catch statistical anomalies. End-users typically do not have interest in all of the detected outliers, but only those relevant to their application. Given an existing black-box sequential anomaly detector, this paper proposes a method to improve its user relevancy using a small number of human feedback. As our first contribution, the method is agnostic to the detector: it only assumes access to its anomaly scores, without requirement on any additional information inside it. Inspired by a fact that anomalies are of different types, our approach identifies these types and utilizes user feedback to assign relevancy to types. This relevancy score, as our second contribution, is used to adjust the subsequent anomaly selection process. Empirical results on synthetic and real-world datasets show that our approach yields significant improvements on precision and recall over a range of anomaly detectors.

%% file: introduction.tex
Sequential anomaly detection is widely used in various fields, including cyber-security \cite{cyber}, scientific discovery\cite{discovery}, operational metrics monitoring\cite{operational}, etc. The goal is to detect unlikely and rare events that are both statistically abnormal and are relevant to user’s specific application. 

Most anomaly detectors provides a ranked list of statistical anomalies. End-users are typically not interested in all statistical anomalies but only in those that are relevant to their specific application \cite{ExpertFeedback_AAD}. As a result, extra manual efforts are needed to locate the subset of interest. Sometimes users need to expand the subset by finding more points with similar characteristics, which can be tedious. One simple example is monitoring the time series of revenue: although sharp increase anomalies and decrease anomalies are all statistical anomalies, a risk management professional may only care about unexpected decreases in revenue. This phenomenon, which is referred to as the “relevancy challenge” in this paper, lies in the center of a successful anomaly detection system for real world problems.

We tackle the relevancy challenge by collecting a small number of feedback from end users. Users can choose to provide feedback on any previously reported anomaly. In our setting, feedback could either be “positive”, if an anomaly is relevant to their application, or “negative” otherwise. The feedback thus collected is used to adjust future anomaly selection process in order to improve relevancy. 

One important assumption in this work is that the \emph{anomaly detector is a black-box}. Only anomaly scores are observed. This is motivated by a fact that, in many scenarios, improving the statistical anomaly detector is orthogonal to improving the relevancy of reported anomalies, since the user's relevance function is domain specific while the definition of generic statistical anomaly is not. Decoupling these two development processes is critical for generalization and customization of an end-to-end system.  To the best of our knowledge, this paper is the first to assume a given black-box statistical anomaly detector, and tries to improve its relevancy using limited user feedback.


Motivated by real world applications, we assume user feedback are generated in \emph{sequential batches}. Within each batch, anomalies are detected for the users to provide feedback. Note that such \emph{sequential batch} setting is different from \emph{pure online} anomaly detection setting where detection happens in real-time for each data point. In comparison, the \emph{sequential batch} setting generates detection for one batch at a time. We argue that sequential batch setting is a more general setting for feedback based anomaly detection system, especially for users who monitor a system and provide feedback on a certain frequency e.g. daily/hourly basis.


Our method is inspired by a simple observation: statistical anomalies are of different types \cite{WaterAnalytics}, only a subset is relevant to user's application. It starts with \textbf{first} identifying the types using unsupervised clustering. \textbf{Second}, each type (or cluster) is assigned a relevancy score, according to the user feedback previously reported within that cluster. \textbf{third}, we prioritize statistical anomalies from high relevancy cluster and de-prioritize the low relevancy ones during future anomaly selection.

The proposed framework is evaluated on a synthetic data set as well as a real world anomaly detection challenge. We document that our approach consistently improves accuracy of four different base detectors.

The remainder of this article is organized as follows. Section 2 reviews related work. Our main approach is discussed in Section 3. Section 4 describes the data used in our work, followed by the experiment results and case reviews in Section 5. Section 6 concludes.

%% file: related_work.tex
While the paper belongs to the broad fields of active learning or semi-supervised learning, we narrow our review of related works to the sub-field of active anomaly detection (AAD).

A common setup in AAD is to tightly couple a specific anomaly detector method with an online optimization algorithm. \citet{FG_AD_OnlineOptimization} is tied with isolation forest\citet{isolation_forest}, whose anomaly score is a weighted leaf node value. It uses online optimization to update the weight to reflect the feedback. \cite{feedback_tree_based} is specifically tied with tree-based methods. \citet{ExpertFeedback_AAD} is tied with the Loda algorithm \cite{loda} . \cite{CyberAttacks_AD_Expert} is also tied with isolation forest.

There are also works on active learning for \emph{online} anomaly detection.  \citet{AD_video} provides a framework for anomaly detection in video analytics field. \citet{sequential_AD_noise_limited_feedback} and  \citet{online_AD_social_network} incorporates expert feedback for dynamic threshold control. They employ a two step approach: a filtering step, which assigns log-likelihoods to each observation, followed by a hedging step, which sets a dynamic threshold of the anomaly. While dynamic threshold control is another important topic for anomaly detection, it is relatively orthogonal to learning user preference over different anomaly types, which is the main focus of this paper. 

\citet{WaterAnalytics} formulates this problem with a semi-supervised clustering approach. It uses COP K-means with cannot-link constraint to incorporate user feedback. The final anomaly score is a function of clustering results including cluster size and each point's distance with labeled anomalies. Thus, it is also tied with its specific anomaly detection algorithm, lacking the capacity of handling generic black-box detector.

%% file: method3.tex
Assume we have a time-series \( V^t=( v_1, v_2, ...v_t ) \), a black-box base anomaly detector generates scores \( P^t = (p_1, p_2, ..., p_t) \) where \(p_i \in [0,1) \), with its detection threshold $\tau_{\anomaly} \in (0,1)$ such that observations with $p_i > \tau_{\anomaly}$ are classified as statistical anomalies.

Detected anomalies are shown to the end-user who then provides positive or negative feedback on a subset of them. We denote \(V^t_{\pos}\) to the set of positively reviewed points, and \(V^t_{\negs}\) the set of negatively reviewed ones. 

\subsection{CLUSTERING}

The first step of our framework is clustering time points with high $p_i$.

\textbf{Candidate Selection}: we define the clustering candidate population by giving certain threshold $\tau^{\candidate} \in [0, 1]$: 

\begin{equation}
    S^{\candidate} := \{p_i | \forall i\in [1,2,..,t]: p_i > \tau^{c }\}.
\end{equation}

Recall $p_i$ denotes the anomaly scores. If \( \tau^{\candidate} = 0 \), all the points are included in the clustering process.

\textbf{Context Collection.} Consider the original time point $v_t$ at time $t$, we define the context vector with length $m$ at time $t$ as
\begin{equation}
    h_t := (v_{t-m},...,v_t).
\end{equation}

We can collect the context set $S^{\context}$ of anomaly candidate set $S^{\candidate}$: 
\begin{equation}
    S^{\context} := \{h_t| \forall z_t \in S^{\candidate} \}.
\end{equation}

\textbf{Sub-sequence Embedding.} We create a low-dimensional embedding of sub-sequence contexts in $S^{\context}$, such that we could obtain the contextual feature map of each time point:
\begin{equation}
    S^{\embedding} := \{\vec{e_t}=f_{\embedding}(h_t) | h_t \in S^{\context}\}.
\end{equation}

\begin{figure}[t]
\vskip 0.2in
\begin{center}
\centerline{\includegraphics[width=0.8\columnwidth]{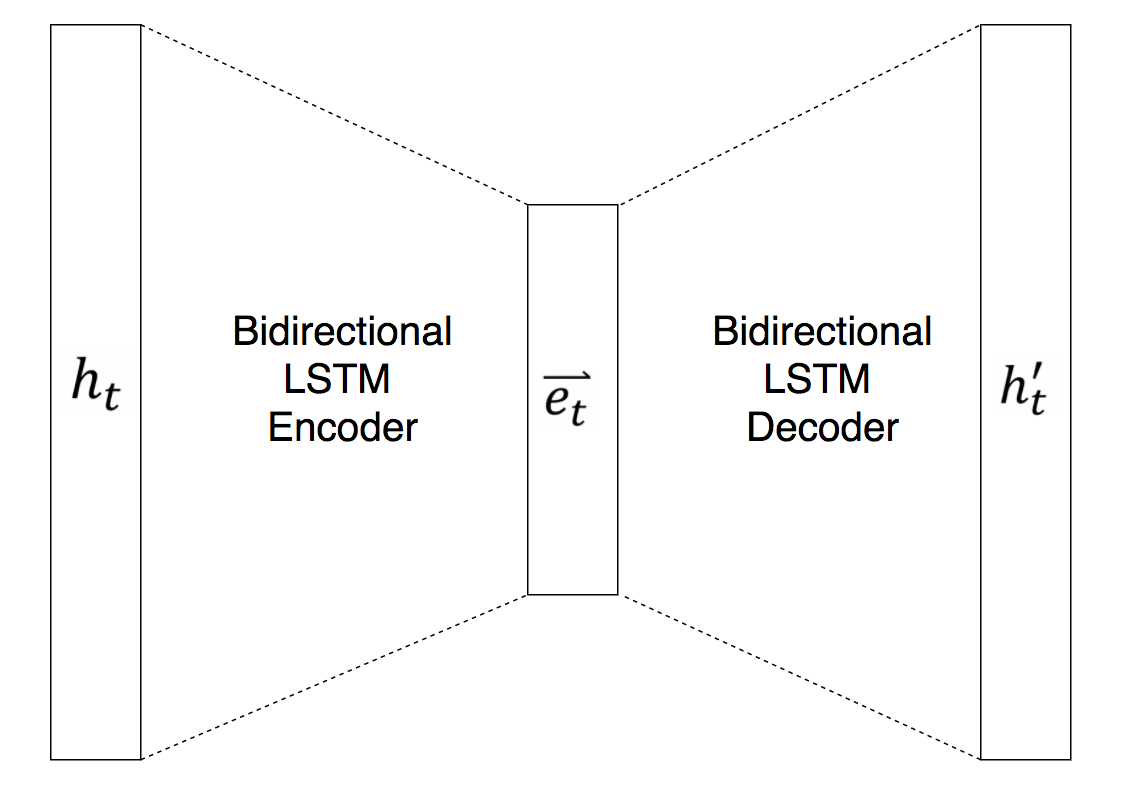}}
\caption{Our sequence embedding module: auto-encoder with Bi-directional LSTM as encoder and decoder}
\label{icml-historical}
\end{center}
\vskip -0.15in
\end{figure}

Here \(f_{\embedding} \) is a sequence embedding function. In this work, we use a sequence auto-encoder with bi-directional LSTM as encoder and decoder, illustrated in Fig \ref{icml-historical}. 
We argue that auto-encoder is very suitable for this task because it can extract vector representation of a sequence efficiently in a totally unsupervised fashion.

\medskip

\textbf{Clustering.} With the feature map $S^{\embedding}$ generated above, we can apply clustering algorithms on top of it:
\begin{equation}
    S^{\cluster} := \{ c_i = g(\vec{e_i}) , c_i \in [1,..,k]| \vec{e_i} \in S^{\embedding}\}
\end{equation}

\(g\) is our clustering function. \(k\) is number of clusters. Specifically, we choose K-means in our framework, thus \(k\) is a hyper-parameter. We provide rationale on this choice in latter chapters.

\subsection{CLUSTER RELEVANCY}
The second main step of our approach is to assign each cluster a relevancy score based on the concentration of feedback and representing the degree to which each cluster is relevant to the users' preferences.

\textbf{Cluster distribution.} 
Consider the distribution $\vec{d}_c$ over clusters on anomaly candidate set $S^{\candidate}$, the distribution $\vec{d}^{\pos}$ over clusters on intersection of $S^\candidate$ and positive labels and the distribution $\vec{d}^\negs$ on intersection of $S^\candidate$ and negative labels. 


\textbf{Cluster relevancy.} Cluster relevancy vector is a function of the three distribution vectors above. It is formulated as:
\[\vec{r} = \frac{exp((\vec{d}^{\pos}-\vec{d}^{\candidate})/\vec{d}^{\candidate})}
                 {exp((\vec{d}^{\negs}-\vec{d}^{\candidate})/\vec{d}^{\candidate})} = exp(\frac{\vec{d}^{\pos} - \vec{d}^{\negs}}{\vec{d}_c})
\]
Intuitively, the numerator represents positive feedback's \emph{distribution deviation} from the overall population, while the denominator is that of negative feedback. The exponential transformation is used for smoothing purposes, since \(\vec{r}\) will be used for scaling in the following steps. In practice, we usually set an upper and lower bound (U, L) on \(\vec{r}\) for regularization.

\subsection{ANOMALY SELECTION ADJUSTMENT}
In the last step, the relevancy score \(\vec{r}\) is used to adjust the anomaly selection process in next incoming batch. 

Suppose next batch's data comes in as 
\[ V^{t:T}=( v_t, v_{t+1}, ...v_T ), \]
with base detector scores 
\[ P^{t:T}=( p_t, p_{t+1}, ...p_T ). \]

Following a similar process as section 3.1,  we conduct inference step on new data batch:
\begin{itemize}
    \item Candidate selection \(S^{\candidate}_{t:T}\)
    \item Context collection \(S^{\context}_{t:T}\).
    \item Apply learned embedding function $f_{\embedding}^{l}$ to get embedding \(S^{\embedding}_{t:T}\).
    \item Apply learned clustering function $g$ to get cluster assignment \(S^{\cluster}_{t:T}\).
\end{itemize}

The base anomaly detector will classify all the observations $S^{b}_{t:T}$ whose scores are larger than arbitrary threshold \(\tau^{\anomaly}\) as anomalies: 

 \[S^{b}_{t:T} := \{ v_i | \forall{p_i} > \tau^{\anomaly} 
 , i = t,t+1,...,T \}.
 \]

We count their occurrence across clusters $S^{\cluster}_{t:T}$:
\[\vec{N}^{b} = [n_1, n_2,.., n_k]^T.\]

The \emph{adjusted cluster-wise anomaly size} is then formulated as:
\begin{equation}
    \vec{N}^{adj} = \vec{N}^{b} \odot \vec{r},
\end{equation}
where $\odot$ is point-wise multiplication. 

With each element in \(\vec{N}^{adj}\) rounded down to integer, it represents our \emph{estimated number of relevant anomalies} in each cluster.

We then select the corresponding number of anomalies per cluster according to \(\vec{N}^{adj}\). Inside each individual cluster, ranking is still produced by the base anomaly detector scores. 

One key advantage of this adjustment approach is: \(\vec{r}\) only impacts the number of anomalies per cluster, while the original anomaly ranking inside each cluster is kept. This property, to its maximum extent, preserves the knowledge from the original base detector.

The procedure described is repeated for every new coming batch, as is illustrated in the pseudo-code of Human-In-The-Loop(HITL) Algorithm \ref{alg:HITL}. We show in the following sections that, even with a limited feedback budget, our procedure can help improve black-box anomaly detectors' performance consistently.

\begin{algorithm}[tb]
   \caption{HITL for black-box anomaly detector}
   \label{alg:HITL}
\begin{algorithmic}
   \STATE {\bfseries Input:} Sequential data batch $P^i$, base anomaly detector
   \FOR{$i=1,2,3,...$}
   \STATE Candidate selection $S_i^{\candidate}$
   \STATE Context collection $S_i^{\context}$
   \IF{$i \geqslant 2$}
   \STATE Apply clustering and sub-sequence embedding
   \STATE Anomaly adjustment based on $\vec{r}_{i-1}$
   \ENDIF
   \STATE Collect feedback from end-users
   \STATE Train sub-sequence embedding function $f_{\embedding}^{i}$
   \STATE Train clustering function $g^{i}$
   \STATE Calculate cluster relevancy $\vec{r}_{i}$
   \ENDFOR
\end{algorithmic}
\end{algorithm}


\medskip

%% file: dataset.tex
We test our framework on two datasets: a simulated dataset, and a real world dataset.  

{\bf KPI dataset} is a labeled anomaly detection dataset released by the AIOPS competition \cite{kpi_challenge_data}. It contains a set of KPI metrics (as time-series) with human annotated anomaly labels from companies including Tencent, eBay, Baidu, Alibaba. Example KPI metrics include the order placement rate on certain website or the CPU usage for hosts, etc. There are 29 time-series in the dataset and it contains around 3M data points with 80k labeled true anomalies. We split all the time-series into two parts, the first half is used as training (when applicable) and the second half is used as testing. 

{\bf Synthetic dataset} contains 100 time series, each generated by 200,000 iid gaussian random variables simulating seconds level time series. Around 0.3\% time points are injected anomalies. Half of the anomalies is random sharp increase added to original value while the other half is sharp decrease. However, only the decrease half is labeled as anomaly which simulates user preference. Ideally, a successful anomaly detection system should be able to identify this preference with a small number of feedback.

%% file: experiment.tex
\begin{table*}[t]
\caption{KPI dataset results comparison between base anomaly detector only versus its combination with human-in-the-loop(HITL) framework}
\vskip 0.15in
\setlength{\tabcolsep}{20pt}
\renewcommand{\arraystretch}{1.1}
\centering
\begin{small}
\begin{tabular}{l|rrr|rrr}
\toprule
\textbf{Base Detector} & 
\multicolumn{3}{c|}{\textbf{Base Only}} & 
\multicolumn{3}{c}{\textbf{Base w/ HITL}} \\

& F1 & Precision & Recall & F1 & Precision & Recall\\
\midrule
\textbf{IID} & 0.26 & 0.39 & 0.39 & 0.31 & 0.48 & 0.39  \\
\textbf{HW}  & 0.26 & 0.32 & 0.35 & 0.32 & 0.40 & 0.38  \\
\textbf{RCF} & 0.39 & 0.36 & 0.60 & 0.43 & 0.41 & 0.62  \\
\textbf{RNN} & 0.27 & 0.38 & 0.37 & 0.34 & 0.45 & 0.39  \\

\bottomrule
\end{tabular}
\end{small}
\label{table:stats_kpi}
\end{table*}
\begin{table*}[t]
\caption{Simulated dataset results comparison between base anomaly detector only versus its combination with human-in-the-loop(HITL) framework}
\vskip 0.15in
\setlength{\tabcolsep}{20pt}
\renewcommand{\arraystretch}{1.1}
\centering
\begin{small}
\begin{tabular}{l|rrr|rrr}
\toprule
\textbf{Base Detector} & 
\multicolumn{3}{c|}{\textbf{Base Only}} & 
\multicolumn{3}{c}{\textbf{Base w/ HITL}} \\

& F1 & Precision & Recall & F1 & Precision & Recall\\
\midrule
\textbf{IID} & 0.38 & 0.39 & 0.63 & 0.55 & 0.44 & 0.74  \\
\textbf{HW}  & 0.52	& 0.42 & 0.70 & 0.59 & 0.47 & 0.79  \\
\textbf{RCF} & 0.11	& 0.09 & 0.14 & 0.16 & 0.13 & 0.21  \\
\textbf{RNN} & 0.58	& 0.47 & 0.76 & 0.65 & 0.56	& 0.78  \\

\bottomrule
\end{tabular}
\end{small}
\label{table:stats_sim}
\end{table*}

\subsection{BASE ANOMALY DETECTOR}

\textbf{IID} model is the simplest model which assumes the data point is generated from an \emph{i.i.d.} Gaussian distribution, parameterized by the sample mean and sample variance of the data. The two-sided tail probability of observations on this Gaussian distribution are used to calculate the anomaly score. 

\textbf{Holt-winters} \cite{chatfield1978holt} is an exponential smoothing-based forecasting models where the prediction distribution is used to calculate the tail probability of observations which constitute the anomaly scores. 

\textbf{Recurrent neural network}, or RNN represents the class of models which employs deep learning to generate forecasts. It uses a similar scheme as classical models to calculate the anomaly score based on the tail probability, yet the forecasts are calculated by neural networks. Specifically, in our experiments we choose the DeepAR \cite{salinas2019deepar} architecture.

\textbf{Random Cut Forest} \cite{rcf}, or RCF represents the class of ensemble methods (similar to isolation forest), which estimates the density of data points directly by a forest of random cut trees. Each tree randomly cuts high dimensional data points into sub-spaces. The number of cuts is required to isolate a point is proportional to its estimated density. The depth of a data point in a tree is a measure of the rarity of this point, which yields an anomaly score.

The above four anomaly detectors are typically used in the field of statistical anomaly detection, yet each of them uses a different methodology. We expect they are representative and are able to serve the purpose of ‘black box’ detector to some extent. 

\subsection{HYPER-PARAMETERS}
Silhouette methods \cite{Silhouettes} were used to determine the number of clusters, with the upper limit of five to avoid overfitting in sparse feedback setting. We choose the auto-encoder's LSTM hidden size to be 20. Experiments show that the method is not very sensitive to this parameter.

\subsection{PERFORMANCE}
Our human-in-the-loop(HITL) framework is evaluated with four base anomaly detectors mentioned above. We compare the "base detector only" version, versus "HITL augmented" version. For HITL, only 20 feedback(10 positive, 10 negative) are randomly sampled from the ground truth labels per batch. In KPI dataset, each batch contains 1500(mins) of data which roughly equals to one day's length. After a total of five batches, we stop collecting more feedback and calculate precision, recall, F1 score for both datasets. Results are shown in Table~\ref{table:stats_kpi} and Table~\ref{table:stats_sim}. Consistent improvements are observed across all four methods on both datasets.

\subsection{CASE REVIEWS}
We conducted a qualitative analysis, focused on the learned cluster relevancy \(\vec{r}\). Two major typologies emerge.

In the \textbf{First Type}, clear disparity exists between \(\vec{d}_c\) and \(\vec{d}^{\pos}\), \(\vec{d}^{\negs}\). This indicates strong user preference on some clusters over the others, which makes anomaly selection adjustment necessary. In one example with RCF as base detector:

\(\vec{d}^c = \begin{bmatrix}
       0.1 \\
       0.4 \\
       0.1 \\
       0.2 \\
       0.2 
     \end{bmatrix}
\vec{d}^{\pos} = \begin{bmatrix}
       0 \\
       1.0 \\
       0 \\
       0 \\
       0 
     \end{bmatrix} 
\vec{d}^{\negs} = \begin{bmatrix}
       0.1 \\
       0.2 \\
       0 \\
       0.3 \\
       0.4 
     \end{bmatrix}
\Rightarrow \vec{r} = \begin{bmatrix}
       0.4 \\
       2.0 \\
       0.2 \\
       0.2 \\
       0.1 
     \end{bmatrix}
\)

This means the model manages to learn from the feedback that the second cluster , where all positive feedback is concentrated on, is probably much more important to the user than the others. For this example, F1 improves from 27\%(base only) to 43\%(base w/ HITL) after only 20 feedback. We take a look at cluster 1, and observe that it mainly consists of sharp decreases in low-value region.

\textbf{Second Type} is where \(\vec{d}^{\pos}\), \(\vec{d}^{\negs}\), \(\vec{d}_c\) are similar in value. In this scenario, our model makes relatively small impact on the existing selected anomalies.